\documentclass[conference]{IEEEtran}
\IEEEoverridecommandlockouts
\usepackage{cite}
\usepackage{amsmath,amssymb,amsfonts}
\usepackage{algorithmicx}
\usepackage{algorithm}
\usepackage{algpseudocode}
\usepackage{graphicx}
\usepackage{textcomp}
\usepackage{subcaption}
\usepackage{xcolor}
\usepackage{ulem}
\usepackage [autostyle, english = american]{csquotes}
\MakeOuterQuote{"}
\def\BibTeX{{\rm B\kern-.05em{\sc i\kern-.025em b}\kern-.08em
    T\kern-.1667em\lower.7ex\hbox{E}\kern-.125emX}}
    
\usepackage{todonotes}

\makeatletter
\let\NAT@parse\undefined
\makeatother
\usepackage{hyperref}
\hypersetup{
   colorlinks=true,
    linkcolor= blue,
    allcolors=blue,
    citecolor = blue,
    filecolor=black,      
    urlcolor=blue,
    }

\begin{document}

\title{In Search of a Lost Metric: Human Empowerment as a Pillar of Socially Conscious Navigation\\
}


\author{\IEEEauthorblockN{Vasanth Reddy Baddam$^{*}$}
\IEEEauthorblockA{
\textit{Virginia Tech}\\
Arlington, USA \\
vasanth2608@vt.edu}
\and
\IEEEauthorblockN{Behdad Chalaki$^{\dagger}$}
\IEEEauthorblockA{\textit{Honda Research Institute USA, Inc.} \\
Ann Arbor, USA \\
behdad\_chalaki@honda-ri.com}
\and
\IEEEauthorblockN{Vaishnav Tadiparthi$^{\dagger}$}
\IEEEauthorblockA{\textit{Honda Research Institute USA, Inc.} \\
Ann Arbor, USA \\
vaishnav\_tadiparthi@honda-ri.com}
\and
\IEEEauthorblockN{Hossein Nourkhiz Mahjoub}
\IEEEauthorblockA{\textit{Honda Research Institute USA, Inc.} \\
Ann Arbor, USA \\
hossein\_nourkhizmahjoub@honda-ri.com}
\and
\IEEEauthorblockN{Ehsan Moradi-Pari}
\IEEEauthorblockA{\textit{Honda Research Institute USA, Inc.} \\
Ann Arbor, USA \\
emoradipari@honda-ri.com}
\and
\IEEEauthorblockN{Hoda Eldardiry}
\IEEEauthorblockA{
\textit{Virginia Tech}\\
Blacksburg, USA \\
hdardiry@vt.edu}
\and
\IEEEauthorblockN{Almuatazbellah Boker}
\IEEEauthorblockA{
\textit{Virginia Tech}\\
Arlington, USA \\
boker@vt.edu}
\thanks{\scriptsize$^{*}$Work done while interning at Honda Research Institute USA, Inc.}
\thanks{\scriptsize$^{\dagger}$Both authors contributed equally.}
}

\maketitle

\begin{abstract}
In social robot navigation, traditional metrics like proxemics and behavior naturalness emphasize human comfort and adherence to social norms but often fail to capture an agent's autonomy and adaptability in dynamic environments. This paper introduces human empowerment, an information-theoretic concept that measures a human's ability to influence their future states and observe those changes, as a complementary metric for evaluating social compliance. This metric reveals how robot navigation policies can indirectly impact human empowerment. We present a framework that integrates human empowerment into the evaluation of social performance in navigation tasks. Through numerical simulations, we demonstrate that human empowerment as a metric not only aligns with intuitive social behavior, but also shows statistically significant differences across various robot navigation policies. These results provide a deeper understanding of how different policies affect social compliance, highlighting the potential of human empowerment as a complementary metric for future research in social navigation.
\end{abstract}

\begin{IEEEkeywords}
Human Empowerment; Social Navigation; Social Compliance
\end{IEEEkeywords}

\section{Introduction}
As robots become more integrated into society, expanding beyond warehouse settings where they handled routine tasks, the need for socially compliant behavior in human environments becomes crucial not only for operational success but also for fostering social and technological sustainability. 
In an inclusive community where everyone can thrive, robots must navigate their surroundings with greater awareness of social conformity to the humans, ensuring that their behavior respects all individuals and reduces barriers to accessibility. 
Crowd navigation is one such domain in which research innovations for robots operating in society must be critically evaluated for their impact on society and the environment
\cite{KRUSE20131726,CHARALAMPOUS201785,francis2023principles}. 

Initially, humans were treated as moving obstacles, but this approach led to issues such as the freezing robot problem and oscillatory behaviors, where robots and humans get stuck in a back-and-forth motion \cite{mavrogiannis2023core}.
To address these, more sophisticated methods have been developed, allowing robots to predict human motion in a predict-then-act pipeline \cite{lindemann2023safe} or in a coupled fashion with the robot's control action, enabling counterfactual reasoning \cite{le2024multi}. 
Despite these advances, the performance metrics used to assess social compliance have remained largely unchanged, with current research focusing mostly on factors like the robot's arrival time at a goal or whether it can reach its destination without colliding with humans. 

Social conformity has primarily been evaluated through concepts such as proxemics, which model personal space around humans and assess whether robots violate these spaces \cite{hall1966hidden,pacchierotti2006proxemics}. Some other works have examined how these proxemic boundaries shift as humans move. For instance, rather than modeling personal space as a static ellipse or circle, the space might resemble a cone shape for a moving pedestrian~\cite{samsani2021socially}.  
Other works have focused on encoding specific social navigation rules that humans follow into robot behavior. These include rules for passing, crossing, or overtaking pedestrians \cite{chen2017socially}, as well as guidelines for how robots should interact with groups of people  or queues of humans \cite{katyal2022learning}. Readers can refer to \cite{gao2022evaluation} for a comprehensive survey of existing metrics.

However, even with these refinements, existing metrics fail to fully measure whether robot movement is socially compliant, as they either try to encode certain behaviors or address this problem in some specific scenarios. 
To address this gap, we propose a new metric based on the concept of empowerment, which has been mathematically defined in the context of robotics and reinforcement learning. 
Empowerment relates to an agent's ability to influence its environment in a way that it can observe. 
In this work, we use the information-theoretic concept of Human Empowerment to assess social compliance in crowd navigation, offering a novel way to evaluate robot behavior in human environments. 
Using human empowerment, we measure how robot behavior indirectly impacts human sense of agency, specifically assessing how their ability to influence the environment changes. 
In our setup, human ability is restricted to their motion, with observations represented through an occupancy grid map relative to them. 

The social behavior of robots is a broad and interdisciplinary topic, covering aspects such as the robot's appearance, user interfaces, and more. 
In this work, however, we focus exclusively on the social navigation problem and robot mobility, particularly how they navigate within human environments.

In this paper, we introduce a novel approach to evaluating social compliance in crowd navigation by leveraging the concept of human empowerment, an underexplored metric in this context. Our key contributions are as follows:
\begin{itemize}
    \item \textbf{Human Empowerment as a Complementary Metric:} While human empowerment has been mathematically defined in prior work, we propose its potential use as a complementary metric to assess social compliance in crowd navigation. This approach addresses limitations in existing methods that primarily rely on proxemics, which have been stated to be insufficient for comprehensively capturing social behavior in dynamic environments \cite{singamaneni2024survey}.
    \item \textbf{Framework for Integrating Human Empowerment:} We provide a framework that integrates human empowerment into the evaluation of social performance of navigation policies. 
    This serves as a starting point for further research, particularly for human factors researchers interested in the intersection of autonomy and social compliance in robot navigation.
    \item \textbf{Statistical Validation of Human Empowerment:} Through numerical simulations, we show that human empowerment offers insights that align intuitively with social behavior but also reveals statistically significant differences across various robot navigation policies. This highlights the robustness of empowerment as a metric for capturing the nuanced effects of different policies on human autonomy.
\end{itemize}

By employing empowerment as a metric of performance for social navigation, our work bridges the gap between information-theoretic approaches and the need for socially compliant robot behavior, thus advancing the field of human-robot interaction in dynamic and shared environments.

The rest of the paper is structured as follows: Section \ref{sec:lit} reviews related work. Section \ref{sec:problem} outlines the problem formulation, and Section \ref{sec:methodoly} presents the methodology including key aspects of human empowerment. Section \ref{sec:results} presents and analyzes the evaluation results, while Section \ref{sec:Stat} describes a statistical analysis of different navigation policies on the proposed metric. 
Finally, Section \ref{sec:conclusion} discusses the key findings and concludes the paper.

\section{Related Work} \label{sec:lit}

\subsection{Social Compliance Metrics in Crowd Navigation}
In crowd navigation for robots, several metrics have been developed to assess social conformity, focusing on how well robots adhere to social norms and interact safely with human pedestrians. 
One such fundamental metric involves proxemics, which evaluates the robot’s respect for personal space by measuring distances maintained between the robot and humans during navigation \cite{pacchierotti2006proxemics}. 
Another key metric is compliance with the Social Force Model (SFM), which compares the robot’s movements to the predicted forces that humans experience, ensuring minimal disruption in pedestrian dynamics \cite{helbing1995social}.

Additionally, pedestrian flow disruption metrics evaluate the extent to which the robot impacts the natural movement of pedestrians, with an emphasis on minimizing trajectory deviations and speed reductions caused by the robot \cite{ferrer2013robot_flow}. 
Human comfort and acceptance are also critical metrics, often assessed through user studies and subjective feedback on how comfortable and safe pedestrians feel around the robot \cite{walters2008human}. 
Goal alignment and social navigation metrics assess how well the robot follows social norms, such as walking on the correct side of a hallway \cite{kruse2013social_nav}, while trajectory predictability measures how naturally human observers can anticipate the robot’s future movements \cite{sisbot2007predictable_robot}.
Finally, interaction time is used to evaluate the efficiency of robot-human encounters, with shorter times indicating smoother,  socially compliant behavior \cite{truong2019interaction_time}. 

{However, it remains unclear what navigation behavior represents social optimality \cite{gao2022evaluation}, and it calls into question whether recreating an exact copy of human motion datasets for behavior naturalness is the ultimate goal.
This leads us to an alternate approach of evaluating planning methods developed for social navigation - empowerment. 
It prioritizes an agent's autonomy by measuring its ability to influence its own environment and maintain control over its future actions. 
}

\subsection{Empowerment}
Early research connecting information theory to human-robot interaction (HRI) through the concept of empowerment provided foundational insights into the control and adaptability of autonomous systems. 
One of the earliest works by Klyubin, Polani, and Nehaniv \cite{klyubin2005empowerment} introduced empowerment as a universal, agent-centric measure of control derived from information theory. 
Formally, empowerment is defined as the maximum potential for an agent to influence its environment through its actions, quantifiable via Shannon's mutual information between the agent’s actions and its future states. 
This concept laid the groundwork for how robots could use empowerment as a guiding principle to autonomously interact with their surroundings while maximizing their ability to influence future states, which is crucial for adaptive human-robot collaboration.

Building on this, Salge, Polani, and others explored empowerment as an intrinsic motivation for robots in dynamic environments, particularly focusing on sensorimotor systems. In their 2012 study, Salge et al. \cite{salge2012keep} demonstrated that maximizing empowerment enables robots to "keep their options open," allowing greater flexibility and responsiveness in HRI contexts. 
This was further elaborated in Salge and Polani’s 2013 survey \cite{salge2014empowerment}, where empowerment was applied as an intrinsic motivation in various empirical studies, showing how robots can maintain meaningful control over interactions with humans by prioritizing flexibility and autonomy. 
These early works provided a solid theoretical and empirical foundation for using empowerment to specify key robotics principles \cite{salge2017empowerment}.

Despite rapid developments in the fields of robot navigation and human-robot interactions, the application of empowerment in social navigation remains under-explored \cite{van2020social}.
The use of empowerment as an agent-centric measure could help robots balance autonomy and social conformity by optimizing their ability to navigate dynamic human environments without sacrificing control over their own future actions.

\section{Problem Formulation}
\label{sec:problem}

\begin{figure}[tb!]
\centering
\includegraphics[width=0.45\linewidth, bb = 250 140 550 480, clip=true]{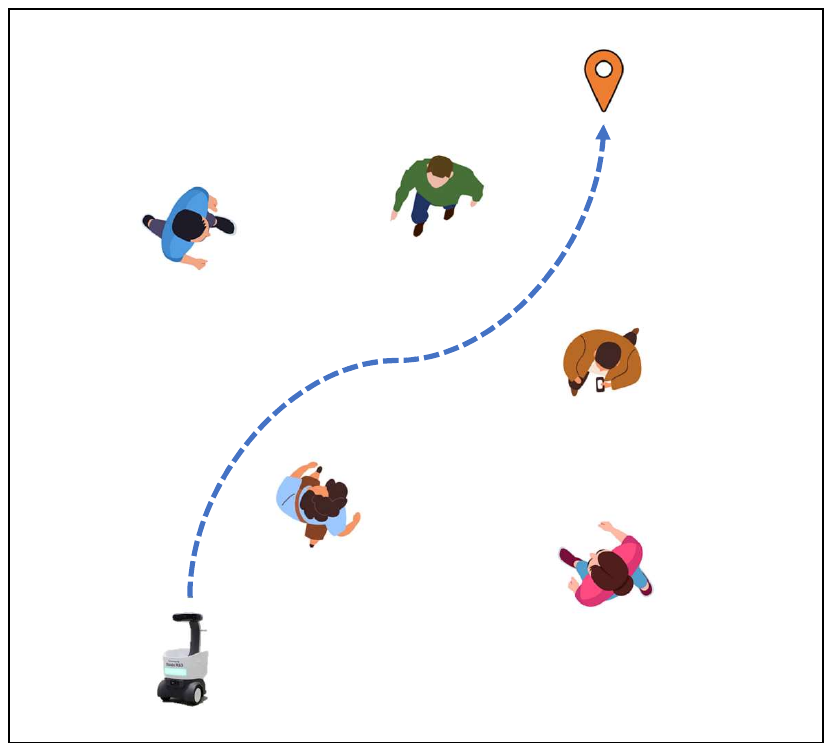}
\caption{A robot navigating in a crowded environment.}
\label{fig:example}
\end{figure}

In this study, we investigate a navigation scenario within a 2-D environment $\mathcal{W} \subset \mathbb{R}^2$, where a single robot operates among $N \in \mathbb{R}$ human pedestrians, as illustrated in Fig.~\ref{fig:example}.
We let $\mathcal{H} = \{1, \ldots, N\}$ represent the set of humans in the scene.
We define the position and velocity of any agent, robot or human, using Cartesian coordinates as 
$\mathbf{s}_{k} = [z_{k}^x, s_{k}^y]^\top \in \mathcal{W}$ and $\mathbf{v}_{k} = [v_{k}^x, v_{k}^y]^\top \in \mathbb{R}^2$, respectively, with $k \in \mathbb{N}$ representing the discrete time-step and $x, y$ the axis components. 
The state of the robot and its control action at time-step $k$ are denoted by $\mathbf{x}_{k} = [\mathbf{s}_{k}^\top, \mathbf{v}_{k}^\top]$ and $\mathbf{u}_{k} = \mathbf{v}_{k}$, respectively. 
A superscript $R$ or $H$ is added below to indicate whether the state of interest refers to that of a robot or a human respectively.

The general objective of social robot navigation is to 
efficiently navigate the robot from its initial position to its goal location within a time limit. 
During navigation, the robot must avoid colliding with other humans while minimizing disruptions to their intended trajectories, which can cause discomfort. This paper focuses exclusively on motion planning, assuming the robot is equipped with onboard sensors that provide the real-time positions of surrounding human agents in the vicinity.

\section{Methodology} \label{sec:methodoly}

\subsection{Mutual Information}
\textbf{Mutual Information (MI)} is a key concept in information theory that quantifies the amount of information shared between two random variables. It measures how much knowing the value of one variable reduces the uncertainty about the other, making it a versatile metric for assessing both linear and nonlinear dependencies.

For two random variables \( X \) and \( Y \), mutual information is defined as:
\begin{equation}
    I(X; Y) = \int\int p(x, y) \log \left( \frac{p(x, y)}{p(x) p(y)} \right) dx \, dy
\end{equation}
where \( p(x, y) \) is the joint probability distribution of \( X \) and \( Y \), and \( p(x) \) and \( p(y) \) are the marginal probability distributions of \( X \) and \( Y \), respectively. This formulation shows that MI is derived from the Kullback-Leibler divergence between the joint distribution \( p(x, y) \) and the product of the marginal distributions \( p(x)p(y) \). In other words, MI measures the difference between the actual joint distribution and what the distribution would be if the variables were independent.

{In planning}, MI can be computed between components such as states, actions, and future states, depending on the behavior analyzed. For example, \( I(s_t; a_t) \) measures the information the current state \( s_t \) provides about the action \( a_t \). Similarly, \( I(a_t; s_{t+1}) \) captures the influence of the action \( a_t \) on the resulting future state \( s_{t+1} \), while \( I(s_t, a_t; s_{t+1}) \) evaluates the combined influence of the state-action pair \( (s_t, a_t) \) on the future state \( s_{t+1} \). This provides insights into the agent’s decision-making process and control over the environment, making MI a powerful metric for evaluating how decisions and states impact future outcomes, valuable for analyzing control strategies.

\subsection{Human Empowerment}

Empowerment is a concept rooted in information theory that measures the extent to which an agent can influence its environment. It is defined as the mutual information between the agent’s actions and the resulting future states of the environment. Essentially, it quantifies the agent’s control over its surroundings. Unlike traditional optimization metrics, which often focus on immediate or specific objectives, empowerment emphasizes maintaining the agent's potential to influence future states, promoting behaviors that preserve adaptability and open possibilities for future decisions.

Formally, empowerment for an agent can be expressed as $ E(s) = \max_{\pi} I(a_t; s_{t+1} | s_t),$
where $a_t$ represents the actions the agent can take at time $t$, $s_{t}$ is the current state, and $s_{t+1}$ is the resulting future state and $\pi$ is the agent’s policy. 
Optimization over policies ensures that the agent selects actions that maximize its influence over future states. 
This approach not only enables the agent to achieve its current objectives but also retains flexibility to adapt and make meaningful decisions as the environment evolves.

Following this, \textbf{Human Empowerment} extends the concept of empowerment to human-robot interactions, focusing on quantifying how much a human’s future state is influenced by the actions they take. In this context, human empowerment measures the degree to which a human's actions affect their future states, serving as a metric for assessing how much empowerment is gained throughout a trajectory. Rather than being tied to any specific goal or objective, this metric focuses purely on the extent of influence exerted by a human's actions.

Human empowerment can be expressed as the mutual information between the human’s actions and the resulting future states in the environment:
\begin{equation} \label{humanempowerment}
    E(z_t^H) = \max_{w} I(a_t; z_{t+1}^H | z_{t}^H),
\end{equation}
   where $z_{t}$ is the human’s current state, $z_{t+1}$ is the future state resulting from the action, and $w$ denotes the human’s policy. 
   Section \ref{occupancymaps} describes how we model the human states $z^H$.

In this formulation, human empowerment directly quantifies how much a human is empowered by taking specific actions, capturing the influence on their future states. This metric is particularly relevant for evaluating social norms, as higher empowerment suggests that the human has more control over their future, aligning better with desirable social interactions. 
Consequently, optimizing robot policies to maximize human empowerment ensures that the human retains significant influence over their trajectory, facilitating behaviors that enhance the human’s agency without assuming any specific goals. 
\subsection{Estimating Human Empowerment}

Human empowerment, as defined in~\eqref{humanempowerment}, is expressed through 
the mutual information between a human’s actions and their resulting future states, which is given by:
\begin{align}
    I(a_t; z_{t+1}^H | z_t^H) = &\int \int p(z_{t+1}^H, a_t | z_t^H) \nonumber \\
    &\times \log \left( \frac{p(z_{t+1}^H, a_t | z_t^H)}{p(z_{t+1}^H | z_t^H) p(a_t | z_t^H)} \right) dz_{t+1}^H da,
\end{align}
where \( p(z_{t+1}^H, a_t | z_t^H) \) denotes the joint probability of future state and actions, while \( p(z_{t+1}^H | z_t^H) \) and \( p(a_t | z_t^H) \) are marginal distributions over future states and actions, respectively.

Given the computational intractability of directly calculating this mutual information, we adopt a variational approximation to estimate it. 
The mutual information can be reformulated using KL divergence between the joint distribution \( p(z_{t+1}^H, a_t | z_t^H) \) and the product of the marginal distributions:
\begin{align}
    I(a_t; &z_{t+1}^H | z_t^H) = \nonumber \\ 
    &D_{KL}(p(z_{t+1}^H, a_t | z_t^H) \parallel p(z_{t+1}^H | z_t^H) \omega(a_t | z_t^H)),
\end{align}
where \( \omega(a_t | z_t^H) \) represents the human's policy and \( p(z_{t+1}^H | z_t^H) \) models the future state distribution given the current state.

\begin{figure}[!h]
    \centering
    \includegraphics[width=0.9\linewidth]{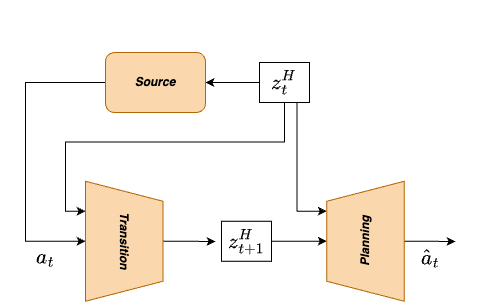}
    \caption{Network modules of Source, Transition, and Planning used to compute Human Empowerment}
    \label{architectureblock}
\end{figure}

To make this estimation tractable, we employ a variational approximation~\cite{mohamed2015variational} and introduce neural networks, each with its own parameterization. 
The pipeline is shown in Fig.~\ref{architectureblock}.

\begin{itemize}
    \item \textbf{Source Policy Network} (\( \omega_{\theta_\omega} \)): This network models the human’s policy \( \omega_{\theta_\omega}(a_t | z_t^H) \), providing the distribution over actions based on the human’s current state. 

    \item \textbf{Transition Network} (\( p_{\theta_T}\)): This network models the environment's state transition dynamics \( p_{\theta_T}(z_{t+1}^H | z_t^H, a_t) \).
    It estimates the future states \( z_{t+1}^H \) 
    conditioned on the current state \( z_t^H \) and action \( a \). It is critical for capturing the influence of a human's actions on their next state.

    \item \textbf{Planning Network} (\( q_{\theta_q} \)): This network provides a variational approximation \( q_{\theta_q}(a_t | z_{t+1}^H) \), modeling the posterior distribution over actions given the future state \( z_{t+1}^H \). 
\end{itemize}

Using these networks, we estimate the mutual information by minimizing the KL divergence between the true state transition distribution \( p_{\theta_T}(z_{t+1}^H | z_t^H, a_t) \) and the variational approximation \( q_{\theta_q}(a_t | z_{t+1}^H) \). This leads to the following lower bound for the mutual information:
\begin{align}
    \hat{I}(a_t; z_{t+1}^H | z_t^H) &= \int \int p_{\theta_T}(z_{t+1}^H, a_t | z_t^H) \nonumber \\
    & \times \log \left( \frac{q_{\theta_q}(a_t | z_{t+1}^H)}{p(a_t | z_t^H)} \right) dz_{t+1}^H da.
\end{align}

In this formulation, \( q_{\theta_q}(a_t | z_{t+1}^H) \) is computed by the Planning Network, while the Transition Network models \( p_{\theta_T}(z_{t+1}^H | z_t^H, a_t) \), and the Source Policy Network models \( \omega_{\theta_\omega}(a_t | z_t^H) \). To compute this efficiently, we minimize the KL divergence between \( p(a_t | z_t^H) \) and \( q_{\theta_q}(a_t | z_{t+1}^H) \), leading to the following optimization objective:
\begin{align} \label{humanempowermentestimate}
    \hat{I}&(a_t; z_{t+1}^H | z_t^H) \approx \nonumber \\ 
    &\mathbb{E}_{p_{\theta_T}(z_{t+1}^H, a_t | z_t^H)} \left[ \log q_{\theta_q}(a_t | z_{t+1}^H) - \log \omega_{\theta_\omega}(a_t | z_t^H) \right].
\end{align}

By optimizing this objective with respect to the network parameters \( \theta_\omega \), \( \theta_T \), and \( \theta_q \), we obtain an efficient estimate of human empowerment, which captures the degree to which a human’s actions influence their future states over the course of a trajectory. We use Monte Carlo sampling to handle the high-dimensional integrals involved in this estimation.

\subsection{Training}
The training process involves updating the Source Policy Network, Transition Network, and Planning Network through gradient-based optimization. Each network is designed to minimize its own loss function, with the goal of accurately estimating the mutual information between actions and future states, which is then used to compute human empowerment. 

The Source Policy Network, parameterized by $ \theta_{\omega} $, models the action distribution $ \omega_{\theta_\omega}(a_t | z_t^H) $, given the current state $ z_t^H $. The loss function for this network is defined as the negative log-likelihood of the observed actions, formulated as 
\begin{equation}
\mathcal{L}_{\text{policy}}(\theta_{\omega}) = - \mathbb{E}_{(z_t^H, a_t) \sim \mathcal{D}} [\log \omega_{\theta_\omega}(a_t | z_t^H)].
\end{equation}
The Transition Network, parameterized by $ \theta_T $, is responsible for predicting the next state $ z_{t+1}^H $ given the current state $ z_t^H $ and action $ a $. Unlike probabilistic models that predict a distribution over possible future states, the Transition Network produces a point estimate of the next state. Therefore, the training objective for this network is to minimize the \textbf{mean square error (MSE)} between the predicted next state and the actual next state. The loss function is expressed as 
\begin{equation}
\mathcal{L}_{\text{transition}}(\theta_T) = \mathbb{E}_{(z_t^H, a_t, z_{t+1}^H) \sim \mathcal{D}} \left[\|z_{t+1}^H - \hat{z}_{t+1}^H\|^2 \right],
\end{equation}
where $ \hat{z}_{t+1}^H $ represents the predicted next state. This ensures that the network captures the transition dynamics accurately and efficiently minimizes the error between the predicted and true states.

The Planning Network, parameterized by $ \theta_q $, approximates the posterior distribution $q_{\theta_q}(a_t| z_{t+1}^H)$, which models the action distribution given the future state. Since the true action distribution $p(a_t| z_t^H)$ is not available, we maximize a variational lower bound (ELBO)~\cite{kingma2013auto} on mutual information, which serves as a tractable approximation. The loss function for the Planning Network includes an entropy regularization term, preventing the predicted action distribution from becoming overly deterministic. The loss function is formulated as 
\begin{align}
\mathcal{L}_{\text{planning}}(\theta_q) = - \mathbb{E}_{q_{\theta_q}(a_t | z_{t+1}^H)} &[\log \omega_{\theta_\omega}(a_t | z_t^H)] \nonumber \\
&+ \lambda \mathbb{H}(q_{\theta_q}(a_t | z_{t+1}^H)),
\end{align}
where $ \lambda $ is a regularization coefficient, and $ \mathbb{H} $ represents the entropy of the action distribution. By maximizing the variational lower bound, the Planning Network ensures that its predictions maintain sufficient diversity and align with the current action distribution predicted by the Source Network.

Training is performed using mini-batch gradient descent. 
At each step, we sample mini-batches of state-action-next state triples $ (z_t^H, a_t, z_{t+1}^H) $ from the dataset $ \mathcal{D} $. The forward pass through the networks computes the respective losses for the Source Policy, Planning, and Transition Networks. Then gradients are computed through backpropagation, and the parameters $ \theta_{\omega}, \theta_T, \theta_q $ are updated iteratively. 
The training process continues until the total loss converges, indicating that the networks have successfully learned to model the human decision-making process, transition dynamics, and action distribution in future states.

Mutual information between actions and future states is estimated by combining the outputs of the Source Policy and Planning Networks, while the Transition Network contributes to predicting the future state based on the current state and action. The mutual information estimate is given by~\eqref{humanempowermentestimate}. We summarize the core pipeline in  Fig.~\ref{architectureblock} and Algorithm~\ref{HEalgorithm}.

\subsection{{Occupancy Maps}} \label{occupancymaps}  
To model human motion and interactions in dynamic environments, we utilize an \textit{occupancy map} to represent the spatial distribution and movement of humans. The occupancy map is a grid-based structure where each cell provides information about the presence and velocity of a neighboring agent. This representation is essential for capturing both static and dynamic elements of the environment, enabling effective navigation and interaction planning.

The occupancy map is formalized as a 3D tensor with dimensions $c \times r \times 3$, where $c$ and $r$ denote the height and width of the grid. Each cell $e_j$ in the grid contains a tuple $e_j = [\text{occupancy}, v_x, v_y]
$:

\begin{itemize}
    \item \textit{occupancy}: A binary value indicating whether the cell is occupied by a human ($1$ for occupied, $0$ for unoccupied).
    \item $v_x, v_y$: The velocity components of the human in the $x$- and $y$-directions, respectively.
\end{itemize}

Each occupancy map is ego-centric, centered around a specific human, and captures the local environment around them. The spatial grid provides an immediate snapshot of the human’s surroundings, with each cell representing a discrete location in the vicinity of the human. The inclusion of velocity components within each cell allows the system to model not only the spatial arrangement of neighboring humans but also their movement patterns, which is crucial for planning future actions in dynamic environments.

We denote the occupancy map around human $k$ as $g_k(i, j) = 
\begin{bmatrix}
\text{occupancy}_{ij} & v_{x,ij} & v_{y,ij}
\end{bmatrix}$: 
Here, $i$ and $j$ represent the row and column indices of the grid, defining the spatial position relative to the human. The complete human state at time $t$, denoted $z_t$, is a concatenation of the individual occupancy maps of all neighboring humans as $z_t^H = [g_1, g_2, \dots, g_k]
$

where each $g_k$ provides a local view of the environment surrounding a particular human. This concatenated structure enables the model to reason about the interactions and potential collisions between multiple humans in the vicinity, facilitating accurate decision-making for navigation tasks.

\begin{algorithm}
\small	
\caption{Human Empowerment}\label{HEalgorithm}
\begin{algorithmic}[1]
\State Initialize Networks : Source Network $(\theta_\omega)$, Transition Network $(\theta_T)$, Planning Network $(\theta_q)$
\State Initialize dataset $\mathcal{D}$
\While{not converged}
\State Sample a batch $<s_t, s_{t+1}>$ from $\mathcal{D}$
\State Compute occupancy map $z^H$ from state $s$
\State Sample action from source policy $a_t \sim \omega_{\theta_\omega} (.| z^{H}_t)$
\State Find future map $z^{H}_{t+1}$ from transition network \Statex $z^{H}_{t+1} = p_{\theta_{T}} (.| z_t^H, a_t)$
\State Compute losses: $\mathcal{L}_{\text{policy}}$, $\mathcal{L}_{\text{transition}}$, and $\mathcal{L}_{\text{planning}}$
    
    \State Update gradients:
    \State $\Delta \theta_\omega \propto \nabla_{\theta_\omega} \log \omega_{\theta_\omega}(a_t | z_t^H)$
    \State $\Delta \theta_T \propto \nabla_{\theta_T} \mathcal{L}_{\text{transition}}$
    \State $\Delta \theta_q \propto \nabla_{\theta_q} \log q_{\theta_q}(a_t | z_{t+1}^H)$
\EndWhile
\State Compute empowerment 
\[E(s)=
    \left[ \log q_{\theta_q}(a_t | z_{t+1}^H) - \log \omega_{\theta_\omega}(a_t | z_t^H) \right]
    .\]

\end{algorithmic}
\end{algorithm}

\section{Experiments and Results}\label{sec:results}

We use a modified version of the CrowdNav simulation environment \cite{chen2019crowd} to evaluate the performance of different social navigation methods on the metric of human empowerment. 
{The environment allows us to control variables difficult to standardize in human studies, such as crowd density and movement patterns. 
Widely used to validate crowd navigation policies, CrowdNav provides a safe, cost-effective, and replicable platform to refine strategies in diverse, simulated scenarios. 
It enables testing of methods based on metrics like collision avoidance, trajectory smoothness, and advanced measures such as discomfort and empowerment. 
This controlled setting facilitates large-scale experimentation, offering insights into robot behavior in human-centric spaces while saving resources and mitigating risks. Promising strategies can be prioritized for subsequent human testing.}

We examine $4$ robot navigation methods in particular - ORCA based on Reciprocal Avoidance \cite{van2011reciprocal} with no safety space, Linear which assumes that the robot has a constant velocity from its origin to its destination, SARL - a value-based Reinforcement Learning (RL) method based on the attention mechanism \cite{chen2019crowd}, and SAC - a Soft Actor-Critic RL algorithm. 
{We select these methods to capture a range of navigation strategies, from basic trajectory optimization to learning-based methods with varying levels of social-awareness, for initial insights into how empowerment varies. 
However, the pipeline can adapt to any policy with available trajectory data.}

We utilize the circle-crossing scenario, where both robots and simulated humans start from a circular arrangement and attempt to cross the circle to reach a diametrically opposite point. 
The simulated humans also use ORCA for their navigation.
This scenario creates a challenging environment as humans and robots converge toward the center, requiring collision avoidance and socially compliant navigation strategies.

We seek primarily to answer the following questions.
\begin{itemize}
    \item Does the empowerment metric validate our intuitions about how humans perceive their autonomy evolving in dynamic social environments?        
    \item Is the metric capable of capturing differences across different navigation policies?
    \item How does the empowerment metric compare to other social conformity metrics, such as discomfort?
\end{itemize}

We believe that from positive responses to these questions, we can demonstrate the utility of empowerment as a metric for evaluating human-robot interactions in social navigation tasks.

\subsection{Temporal Dynamics of Human Empowerment}

\begin{figure}[!ht]
    \centering
    \includegraphics[width=0.99\linewidth]{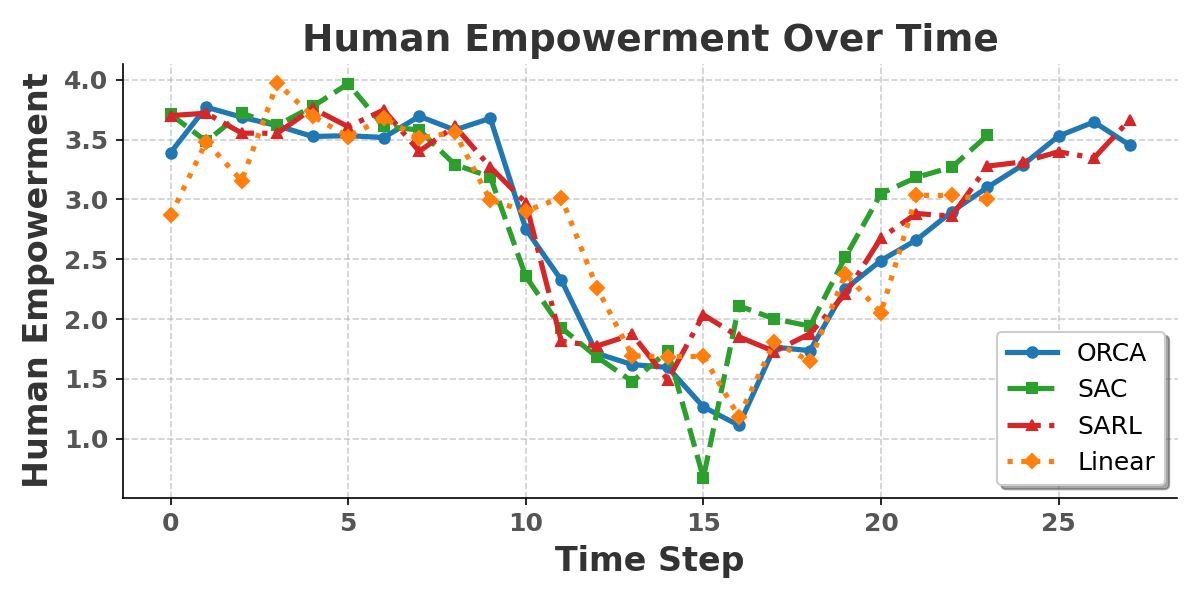}
    \caption{Empowerment for Different Policies vs. Time}
    \label{fig:EmpVsT}
\end{figure}

Fig.~\ref{fig:EmpVsT} illustrates the evolution of human empowerment over time for three navigation methods: ORCA, SAC, and SARL. Initially, empowerment is high, indicating that the human has considerable control over their actions. However, as humans approach closer to each other near the center of the circular environment, empowerment decreases across all methods. This reduction is expected, as proximity to other agents restricts a human’s ability to move freely, diminishing their control and influence over future states. After a few more time steps, empowerment values recover as the humans regain autonomy. This pattern confirms that empowerment effectively captures how an agent's autonomy evolves in response to dynamic changes in social environments.

\begin{figure*}
     \centering
     \begin{subfigure}[b]{0.24\textwidth}
         \centering
         \includegraphics[width=\textwidth]{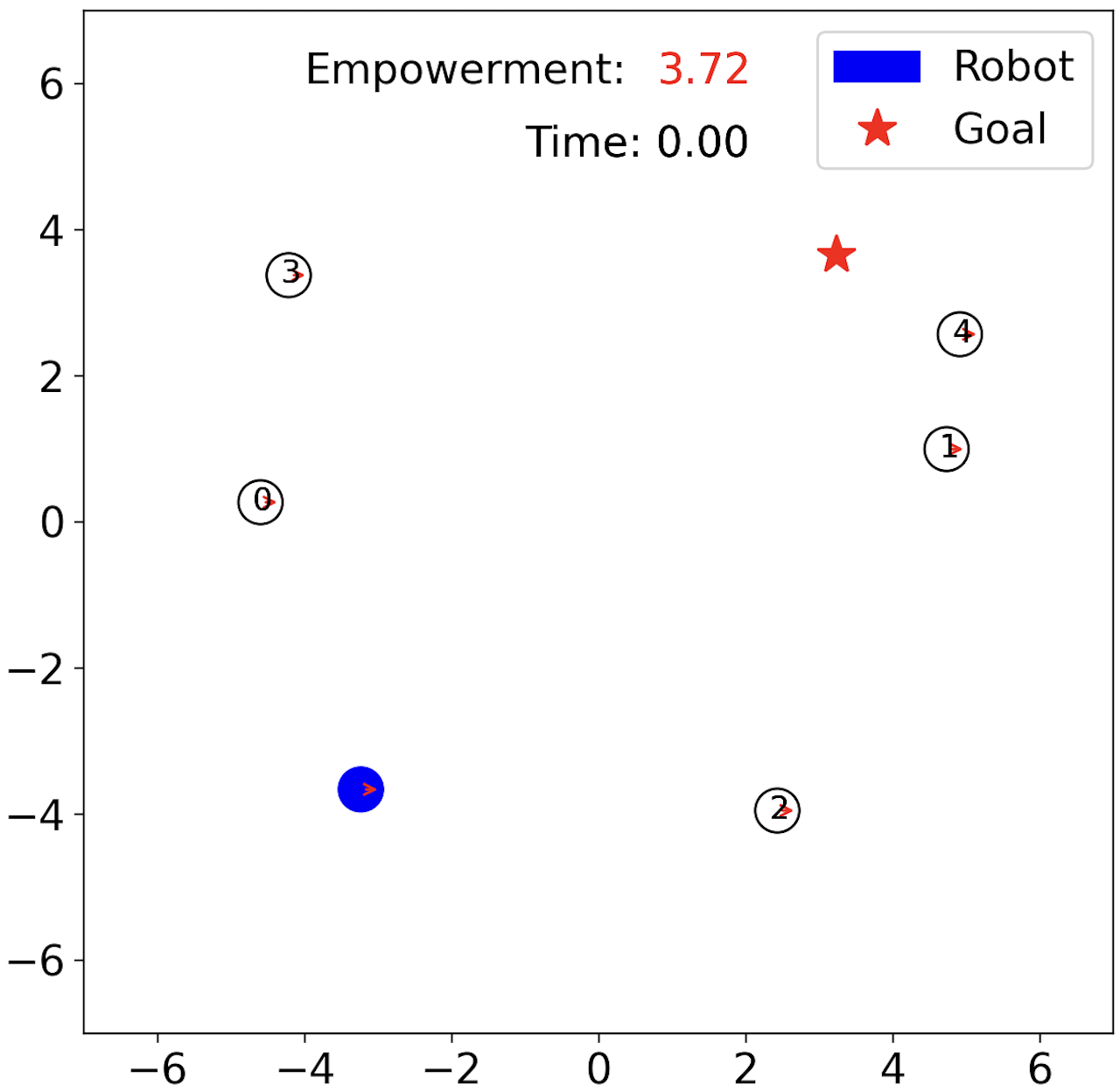}
         \caption{t = 0s}
     \end{subfigure}
     \begin{subfigure}[b]{0.24\textwidth}
         \centering
         \includegraphics[width=\textwidth]{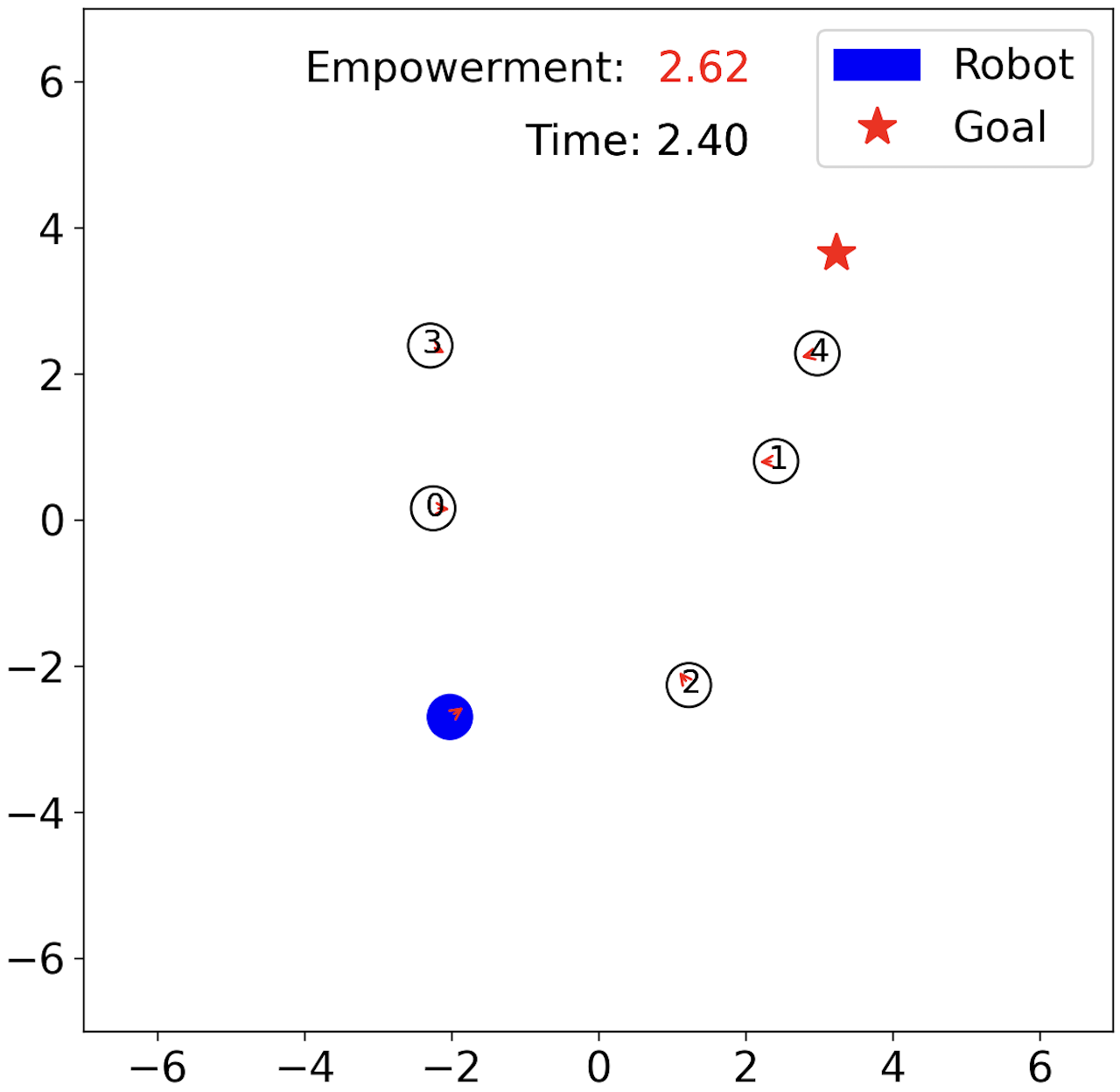}
         \caption{t = 6s}
     \end{subfigure}
     \begin{subfigure}[b]{0.24\textwidth}
         \centering
         \includegraphics[width=\textwidth]{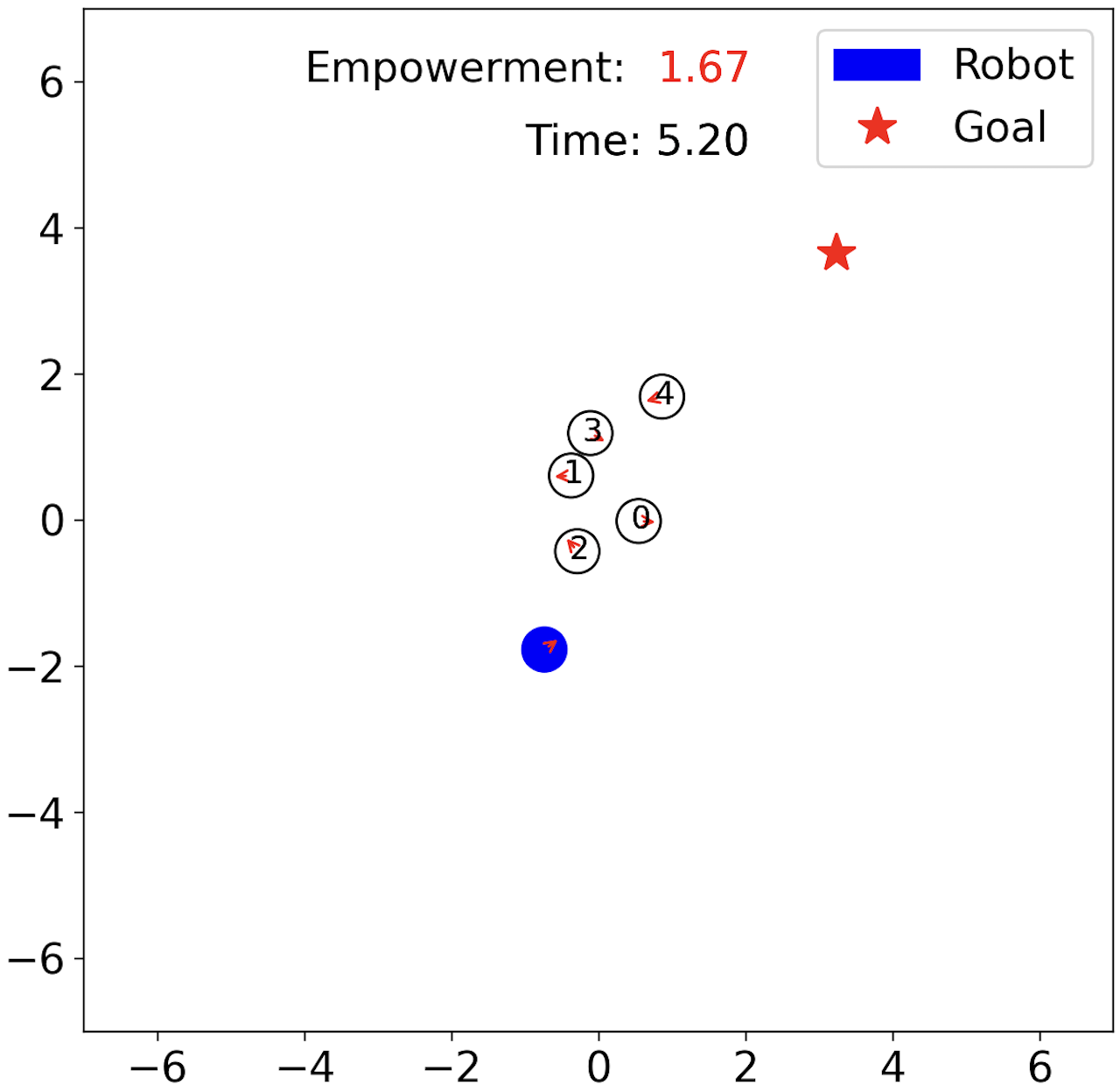}
         \caption{t = 12s}
     \end{subfigure}
     \begin{subfigure}[b]{0.24\textwidth}
         \centering
         \includegraphics[width=\textwidth]{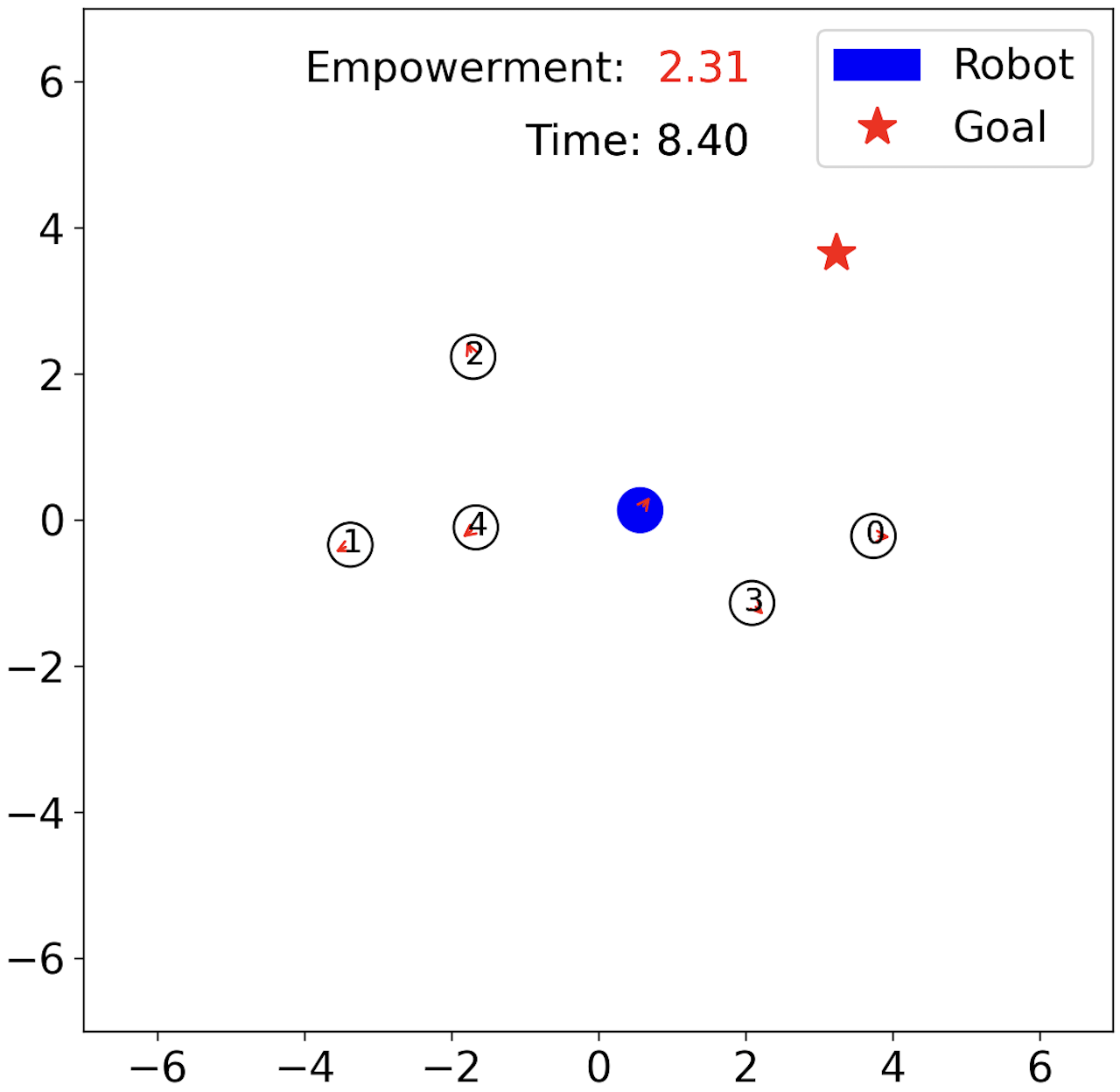}
         \caption{t = 18s}
     \end{subfigure}\\
        \caption{Time-evolution of the robot's trajectory and corresponding empowerment values at different time steps. The robot (blue circle) navigates towards the goal (red star) while interacting with nearby humans (numbered circles). Empowerment, shown in the top of each subplot, decreases as the robot progresses through the environment, reflecting reduced autonomy due to increased proximity to humans. This figure demonstrates how empowerment evolves dynamically in response to the robot's movement through a crowded environment.}
        \label{Trajplots}
\end{figure*}

\subsection{Impact of Crowd Density on Human Empowerment
}

\begin{figure}[!ht]
    \centering
 \includegraphics[width=\linewidth]{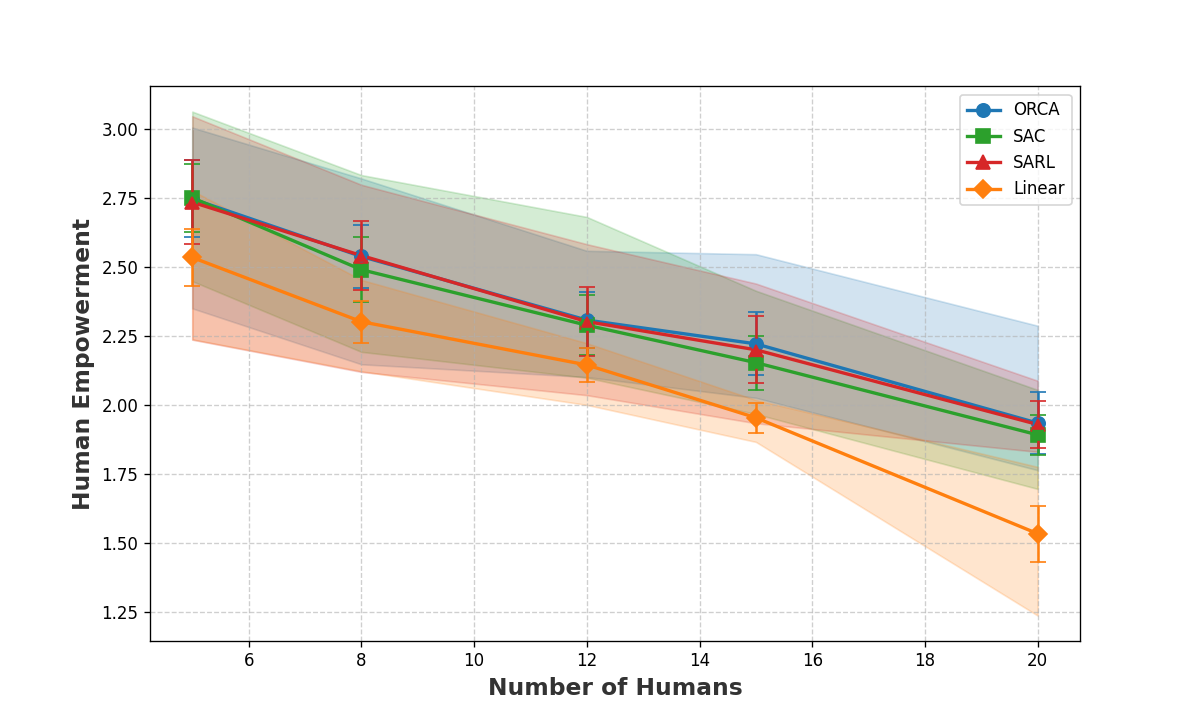}
    \caption{Avg. Empowerment for Different Policies vs. Crowd Sizes}
    \label{fig:EmpVsDen}
\end{figure}

Fig.~\ref{fig:EmpVsDen} shows how human empowerment decreases as the number of humans in the environment increases, across the different navigation methods mentioned previously. 
Initially, with fewer humans, empowerment is relatively high, reflecting the agents' greater autonomy and ability to influence their environment. 
However, as crowd density increases, empowerment steadily declines across all methods, indicating a reduced capacity to act freely in denser crowds. 
This aligns with expectations, as higher crowd densities impose more constraints on human movement, requiring more cautious navigation to avoid collisions and respect personal space. 
The overall downward trend confirms that empowerment effectively captures an agent's reduced freedom to navigate in dense environments, with minor differences between methods showing slight variations in performance.
\subsection{Human Empowerment Across Policies}
\begin{figure}[!ht]
    \centering
    \includegraphics[width=0.95\linewidth]{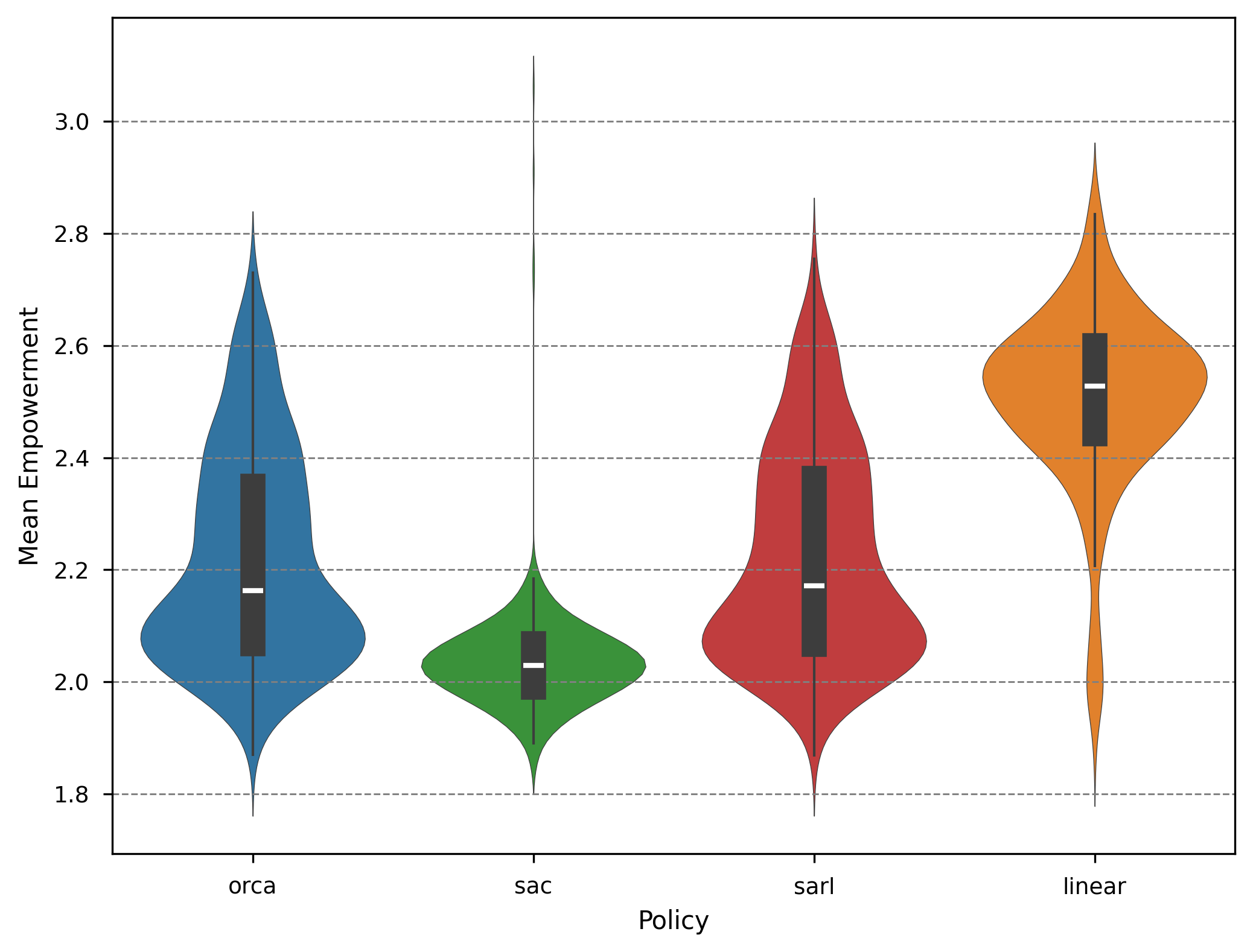}
    \caption{Violin plots of mean empowerment values across $500$ episodes for all policies, showing the distribution and variability for comparison.}
    \label{fig:all_violin}
\end{figure}

We provide violin plots displaying the distribution of mean empowerment values across all trials ($500$ seeds) for each policy, as shown in Fig.~\ref{fig:all_violin}. The figure illustrates the central tendency and spread of empowerment values, allowing for direct comparison between policies. In addition, to provide a clearer understanding of performance in terms of success, we present the success rates for each policy in Table \ref{table:success}. 
Notably, we include all trials (both successful and unsuccessful) in the empowerment analysis. Unsuccessful trials are defined as those where a collision occurred, or the agent failed to reach the goal within the specified time. 
If a collision occurs, the episode ends prematurely. Since we report the mean of average empowerment values over the entire episode and across all humans, early terminations due to collisions may not significantly impact the overall mean empowerment. This explains why scenarios with lower success rates can still exhibit higher empowerment values. Note that these two metrics are independently useful in assessing policy performance.

\begin{table}[!ht]
    \centering
        \caption{Success rate for different policies }
    \begin{tabular}{|c|c|c|c|c|}
    \hline
          Policy & ORCA & SAC & SARL & Linear \\
         \hline
         Success &  $0.456$ & $0.99$ & $0.44$ & $0.05$ \\
         \hline
    \end{tabular}
\label{table:success}
\end{table}

\subsection{Temporal Comparison Between Empowerment and Discomfort Metrics}
\begin{figure}
    \centering
    \includegraphics[width=0.95\linewidth]{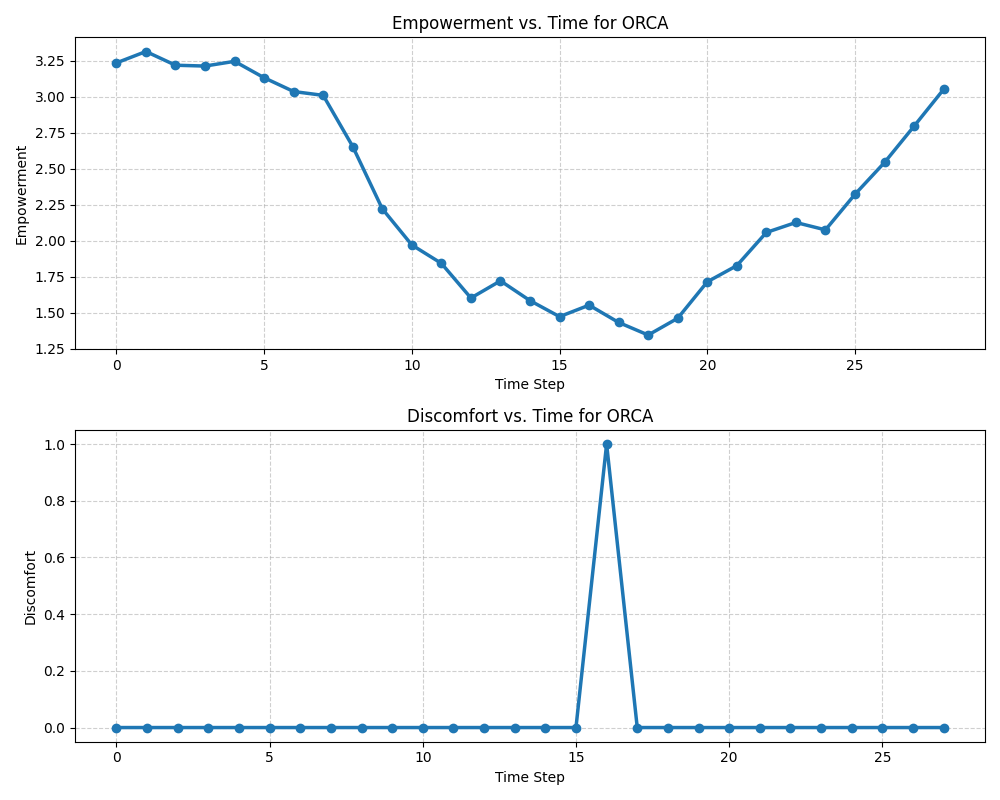}
    \caption{Comparison of Empowerment against Discomfort Rate} 
    \label{fig:EmpVsDisc}
\end{figure}

The two plots in Fig.~\ref{fig:EmpVsDisc} depict empowerment and discomfort over time for the ORCA policy over a single episode, showing an interesting relationship between the two metrics. 
As the top graph indicates, human empowerment decreases steadily as the agents get closer to each other, reflecting their diminishing control and flexibility as they navigate through increasingly constrained conditions. 
Empowerment then begins to recover, rising after some time as the humans regain autonomy.

In contrast, the discomfort metric (bottom graph) remains at zero for most of the timeline but spikes sharply when the humans are close, triggering a discomfort response. 
Clearly, this binary metric is useful only to indicate whether or not a critical event has occurred (e.g., a close encounter) without providing a sense of the gradual changes leading up to that event.
Human empowerment on the other hand, as a continuous variable, offers the benefit of reflecting a more nuanced picture, as it continuously tracks a human's ability to navigate freely. 
This makes empowerment a more informative metric for capturing the dynamics of human-robot interactions.

\section{Statistical Analysis}\label{sec:Stat}
To validate our proposal of using human empowerment as a metric for measuring the performance of a robot's policy, we demonstrate its ability to differentiate the implicit influence of various policies. In this section, we present our statistical analysis to support this objective.

We conduct $500$ trials for each of four different policies, SARL, SAC, ORCA, and Linear.
The objective is to evaluate and compare the impact of these policies on human empowerment. For each trial, we calculate the Mean Empowerment, a key metric for our analysis, and subsequently perform appropriate statistical tests to validate our hypothesis.

In each trial, we compute human empowerment for every individual human at each time step, starting from the beginning of the scenario until the human reached their destination. Then, we compute the average empowerment of each human throughout their trajectory to capture their empowerment throughout the interaction. To derive the final metric, we calculate the average of these empowerment values across all humans in the scene, which we refer to as \textbf{Mean Empowerment} throughout this section. For our statistical analysis, we construct a distribution of $500$ random trials for each policy and record the Mean Empowerment parameter. In each trial, we randomize the human initial position based on random seed that the learning-based policy have not seen during their training. 

We first assess the normality of the collected data using the \textbf{Shapiro-Wilk test} \cite{shapiro1965analysis} due to our sample size ($500$ samples per each policy) and its ability to detect deviations in both skewness and kurtosis. The null-hypothesis for this test is that the data follows a normal distribution. Our result (Table \ref{table:Shapiro}) shows that the mean of average human empowerment is not normally distributed ($p<0.05$), leading us to reject the null-hypothesis.

\begin{table}[!ht]
    \centering
        \caption{Statistical Results for Shapiro-Wilk test }
    \begin{tabular}{|c|c|c|c|c|}
    \hline
         & ORCA & SAC & SARL & Linear \\
         \hline
         Statistic & $0.940$ & $0.632$ & $0.942$ & $0.911$ \\
         \hline
         p-value & $2.5e-13$ & $2.4e-31$ & $5.2e-13$ & $1.5e-16$ \\
         \hline
    \end{tabular}
\label{table:Shapiro}
\end{table}

\begin{figure}[!ht]
    \centering
    \includegraphics[width=0.95\linewidth]{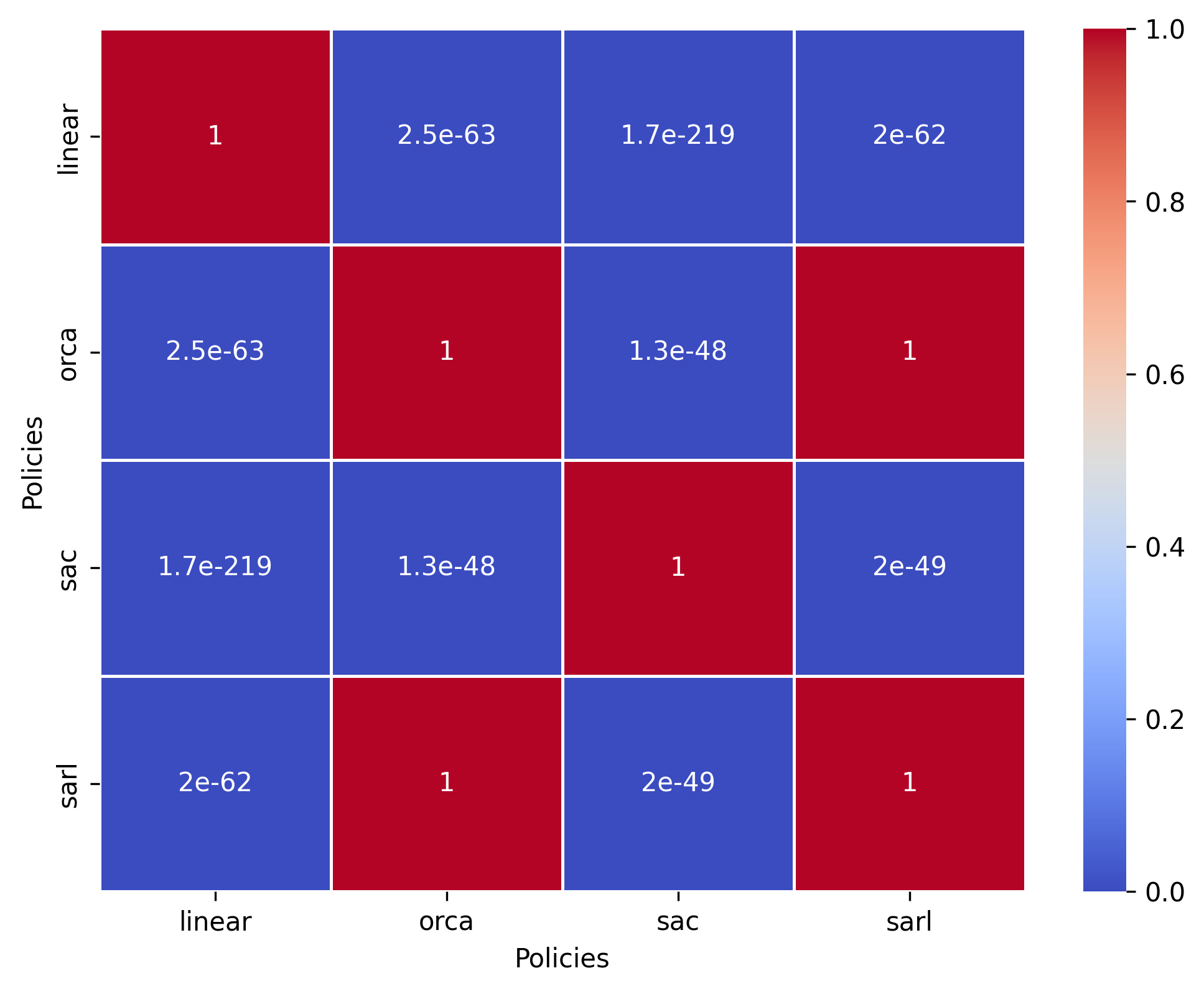}
    \caption{Dunn's post-hoc test showing the \textbf{p-value} between policy pairs.} 
    \label{fig:dunn_test}
\end{figure}

Further, we use a non-parametric \textbf{Kruskal-Wallis test} \cite{kruskal1952use} to determine whether there were statistically significant differences in the empowerment values across the different policies. The null hypothesis for this test states that the distribution of empowerment values is the same across all policies. The test results for this test (Statistics=$1233.4$, p= $8.9e-266$) rejects the null hypothesis and reveals that \textbf{at least one of the policies} has a significantly different distribution compared to the others.
Following the Kruskal-Wallis test, we perform \textbf{a Dunn's post-hoc test} \cite{dunn1964multiple} with Boneferroni correction to identify which specific policy pairs had significant difference between the two policies. The results (Fig. \ref{fig:dunn_test}) show that mean empowerment for SAC and Linear policies are significantly different (p-value $<0.05$) from all the other policies. Clearly, the mean human empowerment metric is capable of distinguishing between the results of different robot navigation policies. 

\section{Conclusion}\label{sec:conclusion}
In this paper, we introduced human empowerment as a novel metric for evaluating social compliance in crowd navigation, addressing the limitations of traditional metrics like proxemics. 
Our framework offers insights into how the metric captures a human's perceived sense of agency while navigating in shared spaces, and our simulations reveal statistically significant differences across various robot navigation policies. 
These findings suggest that human empowerment complements the assessment of how robot navigation policies affect human autonomy in dynamic environments, contributing to the broader understanding of human-robot interaction.

However, we acknowledge that this work is an initial exploration of empowerment as a metric for social compliance. Future research should involve more thorough investigations and user studies to assess whether empowerment-based metrics effectively capture human perceptions of socially compliant robot behavior. Additionally, exploring other forms of empowerment, such as transfer empowerment or robot empowerment, could provide deeper insights into human-robot interactions in social navigation contexts.

Moreover, we believe empowerment-based metrics have potential beyond evaluation and could be used in the behavior design for social navigation. By incorporating these metrics into the design process, future systems could better balance autonomy with social compliance, enhancing the quality of human-robot interaction in complex, shared environments.

\vfill\eject 
\bibliographystyle{IEEEtran.bst} 
\bibliography{References.bib}

\begin{thebibliography}{10}
\providecommand{\url}[1]{#1}
\csname url@samestyle\endcsname
\providecommand{\newblock}{\relax}
\providecommand{\bibinfo}[2]{#2}
\providecommand{\BIBentrySTDinterwordspacing}{\spaceskip=0pt\relax}
\providecommand{\BIBentryALTinterwordstretchfactor}{4}
\providecommand{\BIBentryALTinterwordspacing}{\spaceskip=\fontdimen2\font plus
\BIBentryALTinterwordstretchfactor\fontdimen3\font minus \fontdimen4\font\relax}
\providecommand{\BIBforeignlanguage}[2]{{%
\expandafter\ifx\csname l@#1\endcsname\relax
\typeout{** WARNING: IEEEtran.bst: No hyphenation pattern has been}%
\typeout{** loaded for the language `#1'. Using the pattern for}%
\typeout{** the default language instead.}%
\else
\language=\csname l@#1\endcsname
\fi
#2}}
\providecommand{\BIBdecl}{\relax}
\BIBdecl

\bibitem{KRUSE20131726}
T.~Kruse, A.~K. Pandey, R.~Alami, and A.~Kirsch, ``Human-aware robot navigation: A survey,'' \emph{Robotics and Autonomous Systems}, vol.~61, no.~12, pp. 1726--1743, 2013.

\bibitem{CHARALAMPOUS201785}
K.~Charalampous, I.~Kostavelis, and A.~Gasteratos, ``Recent trends in social aware robot navigation: A survey,'' \emph{Robotics and Autonomous Systems}, vol.~93, pp. 85--104, 2017.

\bibitem{francis2023principles}
A.~Francis, C.~Pérez-D'Arpino, C.~Li, F.~Xia, A.~Alahi, R.~Alami, A.~Bera, A.~Biswas, J.~Biswas, R.~Chandra, H.-T.~L. Chiang, M.~Everett, S.~Ha, J.~Hart, J.~P. How, H.~Karnan, T.-W.~E. Lee, L.~J. Manso, R.~Mirksy, S.~Pirk, P.~T. Singamaneni, P.~Stone, A.~V. Taylor, P.~Trautman, N.~Tsoi, M.~Vázquez, X.~Xiao, P.~Xu, N.~Yokoyama, A.~Toshev, and R.~Martín-Martín, ``Principles and guidelines for evaluating social robot navigation algorithms,'' 2023.

\bibitem{mavrogiannis2023core}
C.~Mavrogiannis, F.~Baldini, A.~Wang, D.~Zhao, P.~Trautman, A.~Steinfeld, and J.~Oh, ``Core challenges of social robot navigation: A survey,'' \emph{ACM Transactions on Human-Robot Interaction}, vol.~12, no.~3, pp. 1--39, 2023.

\bibitem{lindemann2023safe}
L.~Lindemann, M.~Cleaveland, G.~Shim, and G.~J. Pappas, ``Safe planning in dynamic environments using conformal prediction,'' \emph{IEEE Robotics and Automation Letters}, vol.~8, no.~8, pp. 5116--5123, 2023.

\bibitem{le2024multi}
V.-A. Le, V.~Tadiparthi, B.~Chalaki, H.~N. Mahjoub, J.~D’sa, E.~Moradi-Pari, and A.~A. Malikopoulos, ``Multi-robot cooperative navigation in crowds: A game-theoretic learning-based model predictive control approach,'' in \emph{2024 IEEE International Conference on Robotics and Automation (ICRA)}.\hskip 1em plus 0.5em minus 0.4em\relax IEEE, 2024, pp. 4834--4840.

\bibitem{hall1966hidden}
E.~T. Hall, ``The hidden dimension,'' \emph{Garden City}, 1966.

\bibitem{pacchierotti2006proxemics}
E.~Pacchierotti, H.~I. Christensen, and P.~Jensfelt, ``Evaluations of distance for passage for a social robot,'' in \emph{15th Annual IEEE International Symposium on Robot and Human Interactive Communication (RO-MAN06)}, 2006, pp. 315--320.

\bibitem{samsani2021socially}
S.~S. Samsani and M.~S. Muhammad, ``Socially compliant robot navigation in crowded environment by human behavior resemblance using deep reinforcement learning,'' \emph{IEEE Robotics and Automation Letters}, vol.~6, no.~3, pp. 5223--5230, 2021.

\bibitem{chen2017socially}
Y.~F. Chen, M.~Everett, M.~Liu, and J.~P. How, ``Socially aware motion planning with deep reinforcement learning,'' in \emph{2017 IEEE/RSJ International Conference on Intelligent Robots and Systems (IROS)}.\hskip 1em plus 0.5em minus 0.4em\relax IEEE, 2017, pp. 1343--1350.

\bibitem{katyal2022learning}
K.~Katyal, Y.~Gao, J.~Markowitz, S.~Pohland, C.~Rivera, I.-J. Wang, and C.-M. Huang, ``Learning a group-aware policy for robot navigation,'' in \emph{2022 IEEE/RSJ International Conference on Intelligent Robots and Systems (IROS)}.\hskip 1em plus 0.5em minus 0.4em\relax IEEE, 2022, pp. 11\,328--11\,335.

\bibitem{gao2022evaluation}
Y.~Gao and C.-M. Huang, ``Evaluation of socially-aware robot navigation,'' \emph{Frontiers in Robotics and AI}, vol.~8, p. 721317, 2022.

\bibitem{singamaneni2024survey}
P.~T. Singamaneni, P.~Bachiller-Burgos, L.~J. Manso, A.~Garrell, A.~Sanfeliu, A.~Spalanzani, and R.~Alami, ``A survey on socially aware robot navigation: Taxonomy and future challenges,'' \emph{The International Journal of Robotics Research}, vol.~43, no.~10, pp. 1533--1572, 2024.

\bibitem{helbing1995social}
D.~Helbing and P.~Molnar, ``Social force model for pedestrian dynamics,'' \emph{Physical review E}, vol.~51, no.~5, p. 4282, 1995.

\bibitem{ferrer2013robot_flow}
G.~Ferrer, A.~Garrell, and A.~Sanfeliu, ``Robot navigation in urban environments: Socially compliant navigation among pedestrians,'' \emph{Robotics and Autonomous Systems}, vol.~61, no.~12, pp. 1153--1165, 2013.

\bibitem{walters2008human}
M.~L. Walters, D.~S. Syrdal, K.~L. Koay, K.~Dautenhahn, and R.~Te~Boekhorst, ``Human approach distances to a mechanical-looking robot with different robot voice styles,'' in \emph{RO-MAN 2008-The 17th IEEE international symposium on robot and human interactive communication}.\hskip 1em plus 0.5em minus 0.4em\relax IEEE, 2008, pp. 707--712.

\bibitem{kruse2013social_nav}
T.~Kruse, A.~K. Pandey, R.~Alami, and A.~Kirsch, ``Human-aware robot navigation: A survey,'' \emph{Robotics and Autonomous Systems}, vol.~61, no.~12, pp. 1726--1743, 2013.

\bibitem{sisbot2007predictable_robot}
E.~A. Sisbot, L.~F. Marin-Urias, R.~Alami, and T.~Simeon, ``A human aware mobile robot motion planner,'' \emph{IEEE Transactions on Robotics}, vol.~23, no.~5, pp. 874--883, 2007.

\bibitem{truong2019interaction_time}
N.~Truong and T.~Ngo, ``Efficient crowd-robot interaction for human-aware navigation in dense crowds,'' in \emph{Proceedings of the IEEE International Conference on Robotics and Automation (ICRA)}.\hskip 1em plus 0.5em minus 0.4em\relax IEEE, 2019, pp. 1395--1401.

\bibitem{klyubin2005empowerment}
A.~S. Klyubin, D.~Polani, and C.~L. Nehaniv, ``Empowerment: A universal agent-centric measure of control,'' in \emph{2005 IEEE Congress on Evolutionary Computation}, vol.~1.\hskip 1em plus 0.5em minus 0.4em\relax IEEE, 2005, pp. 128--135.

\bibitem{salge2012keep}
C.~Salge, C.~Glackin, and D.~Polani, ``Keep your options open: An information-based driving principle for sensorimotor systems,'' \emph{Artificial Life}, vol.~17, no.~1, pp. 91--115, 2012.

\bibitem{salge2014empowerment}
------, ``Empowerment--an introduction,'' \emph{Guided Self-Organization: Inception}, pp. 67--114, 2014.

\bibitem{salge2017empowerment}
C.~Salge and D.~Polani, ``Empowerment as replacement for the three laws of robotics,'' \emph{Frontiers in Robotics and AI}, vol.~4, p. 260425, 2017.

\bibitem{van2020social}
T.~van~der Heiden, F.~Mirus, and H.~van Hoof, ``Social navigation with human empowerment driven deep reinforcement learning,'' in \emph{Artificial Neural Networks and Machine Learning -- ICANN 2020}.\hskip 1em plus 0.5em minus 0.4em\relax Springer International Publishing, 2020, pp. 395--407.

\bibitem{mohamed2015variational}
S.~Mohamed and D.~J. Rezende, ``Variational information maximisation for intrinsically motivated reinforcement learning,'' in \emph{Proceedings of the 28th International Conference on Neural Information Processing Systems - Volume 2}.\hskip 1em plus 0.5em minus 0.4em\relax MIT Press, 2015, p. 2125–2133.

\bibitem{kingma2013auto}
D.~P. Kingma and M.~Welling, ``{Auto-Encoding Variational Bayes},'' in \emph{2nd International Conference on Learning Representations, {ICLR}}, 2014.

\bibitem{chen2019crowd}
C.~Chen, Y.~Liu, S.~Kreiss, and A.~Alahi, ``Crowd-robot interaction: Crowd-aware robot navigation with attention-based deep reinforcement learning,'' in \emph{2019 international conference on robotics and automation (ICRA)}.\hskip 1em plus 0.5em minus 0.4em\relax IEEE, 2019, pp. 6015--6022.

\bibitem{van2011reciprocal}
J.~Van Den~Berg, S.~J. Guy, M.~Lin, and D.~Manocha, ``Reciprocal n-body collision avoidance,'' in \emph{Robotics Research: The 14th International Symposium ISRR}.\hskip 1em plus 0.5em minus 0.4em\relax Springer, 2011, pp. 3--19.

\bibitem{shapiro1965analysis}
S.~S. Shapiro and M.~B. Wilk, ``An analysis of variance test for normality (complete samples),'' \emph{Biometrika}, vol.~52, no. 3-4, pp. 591--611, 1965.

\bibitem{kruskal1952use}
W.~H. Kruskal and W.~A. Wallis, ``Use of ranks in one-criterion variance analysis,'' \emph{Journal of the American statistical Association}, vol.~47, no. 260, pp. 583--621, 1952.

\bibitem{dunn1964multiple}
O.~J. Dunn, ``Multiple comparisons using rank sums,'' \emph{Technometrics}, vol.~6, no.~3, pp. 241--252, 1964.

\end{thebibliography}
\end{document}


\section{Appendix}

\subsection{Dynamics}
To provide a detailed understanding of the underlying mechanics and decision-making processes used in our model, SAC, we describe the dynamics, action space, and reward functions for the agents (robots and humans) in the environment.
\subsection{Robot Dynamics} \label{robotdynamics}
We use holonomic kinematics to describe the nominal motion of both agents (robot and humans) in the environment as given below:

\begin{align}\label{dynamics}
    p_x' &= p_x + v_x  \Delta t, \\ 
    p_y' &= p_y + v_y  \Delta t,
\end{align} 
where \( p_x, p_y \) are the positions of any agent on the x and y axes, respectively, and \( v_x, v_y \) are their velocities along the x and y axes. The variables \( p_x', p_y' \) represent the updated positions after a time step \( \Delta t \). Let \( \rho \) be the radius of an agent, and \( \Delta t \) be the step time of the environment.

The attributes of the robot for the learning algorithm are defined as follows:

\begin{itemize}
    \item \textbf{Full State (\( s \))}: The full state of robot consists of its position, velocity, radius, assigned goal position, preferred velocity, and heading angle, expressed as \( s = [p_x, p_y, v_x, v_y, \rho, g_x, g_y, v_{ref}, \theta] \).
    \item \textbf{Observable State (\( \tilde{s} \))}: The observable state of each agent consists of its position, velocity, and radius, expressed as \( \tilde{s} = [p_x, p_y, v_x, v_y, \rho] \).
\end{itemize}

We assume that robot \( i \) observes the local states of the nearest \( m \) agents, with its full state given as \( o_i = [s_i, \tilde{s}_{i1}, \cdots, \tilde{s}_{im}] \).

\subsubsection{Action Space}
The action space for each agent is defined as \( a = (v_x, v_y) \), where the velocity components \( v_x \) and \( v_y \) are continuous and range from \([-1, 1]\), i.e., \( -1 \leq v_x, v_y \leq 1 \).

\subsection{Reward Functions}

We use three reward functions to encourage robots to reach their designated goals and to discourage collisions with humans:

\begin{enumerate}
    \item \textbf{Goal reward (\( r_g \))}: This is the negative Euclidean distance between the robot's current position and its assigned goal position, given as \( r_g = -d_g \).
    \item \textbf{Collision reward (\( r_c \))}: Each robot is penalized \( r_c = -20 \) for each collision with other humans.
    \item \textbf{Success reward (\( r_s \))}: Robot receives \( r_s = 20 \) when it reaches its assigned goal. 
\end{enumerate}

The final reward for robot \( r_i \) is given as:
\begin{equation} \label{rewardstructure}
    r_i = \begin{cases}
    r_s, & \text{if } d_g < \rho, \\ 
    r_g + r_c + r_{int}, & \text{otherwise}
    \end{cases}
\end{equation}

\subsection{Soft Actor-Critic (SAC)} \label{sec:sac}

Soft Actor-Critic (SAC) is an off-policy algorithm~\cite{haarnoja2018soft} designed to address the maximum-entropy control problem by finding a policy $\pi$ that solves the problem while maintaining a high level of action entropy whenever possible.
It accomplishes this by augmenting the total expected return with an entropy component, as $ J = \mathbb{E}\left[
    \sum_{t=0}^T \gamma^t
    \left( r_i^t + \alpha H\left( \pi(o_i) \right) \right)
    \right],$ where {$ H(\pi(o_i))=\mathbb{E}_{a \sim \pi(o_i)}[-\log \pi(a_i \mid o_i)]$} is the entropy of policy $\pi$ at observation $o_i$ of a robot $i$, and $\alpha>0$ is a trainable entropy parameter coefficient which determines the relative importance between the expected returns and the entropy-maximization objective.
    In practice, SAC is an off-policy learning algorithm that 
employs a replay buffer containing past transitions {$\mathcal{D}=\left\{\left(o_i, a_i, r_i, o_i^{\prime}\right)_p\right\}_{p=0}^P$}. 
SAC concurrently trains a parametric policy models $\{\pi_\theta^i\}_{i=0}^n$, and two centralized state-action value models $\{Q^i_{\phi_1}\}_{i=0}^n, \{Q^i_{\phi_2}\}_{i=0}^n$ for each robot $i$. 
SAC uses Bellman error for each state-action function and also utilizes the clipped double-Q trick, and takes the minimum values between the two Q networks. 
The Q network is trained to minimize the following loss:

{{
\begin{equation} \label{qloss}
\begin{aligned}
    J_{Q_{\phi_j}} = & \mathbb{E}_{\substack{o_i, a_i, r_i, o_i' \sim \mathcal{D}, \\ \Tilde{a}_i' \sim \pi^i_\theta(.|o_i')}}
    \bigg[
    \left( y(o_i', r_i, \Tilde{a}_i') - Q_{\phi_j}(o_i, a_i) \right)^2
    \bigg],
\end{aligned}
\end{equation}}}
where the target $y$ is given by 

\begin{equation} \label{qtarget}
\begin{aligned}
    y(o_i', r_i, \Tilde{a}_i') &= r_i +\\ &\gamma 
    \left( 
        \min_{j = 1, 2} Q^i_{\bar{\phi}_{j}} (o_i', \Tilde{a}_i') - \alpha_i \log \pi^i_\theta(\Tilde{a}_i' \mid o_i')
    \right).
\end{aligned}
\end{equation}

In \eqref{qtarget}, $Q^i_{\bar{\phi}_{j}}$ is a frozen target model that is updated at a slower pace than $Q^i_{\phi_j}$ to improve stability.
The policy network is trained by maximizing the following:

{\begin{equation} \label{piloss}
    J_{\pi^i_\theta} = \mathbb{E}_{\substack{o_i \sim \mathcal{D}, \\ \Tilde{a}_i \sim \pi^i_\theta(.|o_i)}} \left[ \min_{j = 1, 2} Q^i_{\bar{\phi}_{j}} (o_i', \Tilde{a}_i) - \alpha \log \pi^i_\theta(\Tilde{a}_i \mid o_i) \right].
\end{equation}}
\subsection{Entropy}
The entropy term in $ J = \mathbb{E}\left[
    \sum_{t=0}^T \gamma^t
    \left( r_i^t + \alpha H\left( \pi(o_i) \right) \right)
    \right],$ can be interpreted as an intrinsic reward given to the robot, which is high for high-entropy policies and low for low-entropy policies. Therefore, the robot will seek not only to maximize the entropy of the policy in the visited states but also to visit states associated with a high-entropy policy. The high entropy accounts for the higher exploration, while lower entropy exploits the visited states, and higher entropy, helps prevent the policy from getting stuck in the local optima. 

The hyperparameter $\alpha$ plays a central role in SAC, determining how much high-entropy states are preferred to pure rewards. 
Choosing a reasonable $\alpha$ can be difficult since it is not directly interpretable, and a suitable value depends dynamically on the current policy's expected returns and entropy. ~\cite{haarnoja2018softapplications} proposed to automatically adjust $\alpha$ by minimizing its own objective,
\begin{equation} \label{alphaloss}
J_\alpha =\alpha \mathbb{E}_{o_i \sim \mathcal{D}}\left[-\log \pi^i_\theta(o_i)\right]-\alpha \bar{H},
\end{equation}
where $\bar{H}$ is a given target entropy. 
In practice, \eqref{alphaloss} modulates $\alpha$, increasing if the current policy entropy is lower than the target entropy and vice versa. 
In contrast to choosing a value of $\alpha$, choosing a value of $\bar{H}$ is much simpler since it is broadly interpretable as the logarithm of the number of actions that we want the max-entropy policy to consider in an average state.
\subsection{Imitation Learning}
Imitation learning is a technique used to train robot by leveraging expert demonstrations. We employ imitation learning to enhance the performance of our robots, using the Soft Actor-Critic (SAC) algorithm as the foundation.

\subsubsection{Expert Data Collection}

The first step in our approach is the collection of expert data. We gather trajectories of the robots using ORCA~\cite{van2011reciprocal} as the behavior policy, which serves as high-quality demonstrations of the desired behavior in the crowd navigation environment. These trajectories include sequences of states, actions, next states, next actions, and rewards, which are then stored in a replay buffer. The quality of the expert data is crucial, as it directly influences the effectiveness of the imitation learning process.

\subsubsection{Training the Actor and Critic}

After collecting the expert data, we proceed to the training phase, where we sample the data from the replay buffer to train the actor and critic networks. The training process is divided into two parts:
\paragraph{Actor Training:} 
The actor is trained using the L2 norm, which minimizes the difference between the actions taken by the actor and the expert actions. Specifically, the loss function for the actor is defined as the average L2 norm of the difference between the predicted action \( a \) and the expert action \( a_{\text{e}} \) over a batch of samples:
\begin{equation}
    J_{\pi_{\theta}} = \mathbb{E}_{\substack{{({a_{\text{e}}) \sim \mathcal{D}_{\text{e}}}}, \\
    a \sim \pi_\theta(.|o_{\text{e}}')}}
    \left[ \| a - a_{\text{e}} \|_2^2 \right],
\end{equation}
where $\mathcal{D}_{\text{e}}$ is the expert replay buffer. This approach encourages the actor to imitate the expert by producing actions that are as close as possible to those demonstrated in the expert data.

\paragraph{Critic Training:} 
The critic is trained using the Bellman error, which measures the difference between the predicted Q-values and the target Q-values derived from the Bellman equation. The loss function for the critic is given by:
{{
\begin{equation} \label{qlossimi}
\begin{aligned}
    J_{Q_{\phi_j}} = & \mathbb{E}_{\substack{o_{\text{e}}, a_{\text{e}}, \tilde{a}_{\text{e}}' r_{\text{e}}, o_{\text{e}}' \sim \mathcal{D}_{{\text{e}}}}}
    \bigg[
    \left( y(o_e', r_e, \Tilde{a}_e') - Q_{\phi_j}(o_i, A) \right)^2
    \bigg],
\end{aligned}
\end{equation}}}
where the target $y$ is given by 

\begin{equation} \label{qtargetimi}
\begin{aligned}
    y(o_e', r_e, \Tilde{a}_e') &= r_e + \gamma 
    \left( 
        \min_{j = 1, 2} Q^j_{\bar{\phi}_{j}} ({o_i}_{\text{e}}', \Tilde{a}_e'))
    \right).
\end{aligned}
\end{equation}

This loss encourages the critic to accurately estimate the value of actions in the environment, guiding the actor to improve its policy.

\subsubsection{Integration with Soft Actor-Critic}

By integrating imitation learning with the SAC framework, we leverage the advantages of both approaches. The SAC algorithm, known for its robustness and efficiency in continuous control tasks, is enhanced by the incorporation of expert knowledge through imitation learning. This hybrid approach allows the agent to learn more effectively, particularly in complex environments where exploration alone may be insufficient to achieve optimal performance. 
\bibliographystyle{IEEEtran.bst} 
\bibliography{References.bib}